\documentclass[twoside,11pt]{article}
\usepackage{jair, theapa, rawfonts}
\usepackage[misc]{ifsym}
\usepackage{ulem}
\usepackage{soul}
\usepackage{url}
\usepackage{latexsym}
\usepackage[utf8]{inputenc}
\usepackage{multicol}
\usepackage{caption}
\usepackage{multirow}
\usepackage{graphicx}
\usepackage{amsmath}
\usepackage{booktabs}
\usepackage{algorithm}
\usepackage{algorithmic}
\usepackage{stfloats}
\usepackage{comment}
\usepackage{ulem}
\usepackage{color}
\usepackage{array}
\usepackage{booktabs}
\usepackage{amsfonts}

\begin{document}


\title{Exploiting Contextual Target Attributes for Target Sentiment Classification
}

\author{\name Bowen Xing \email bwxing714@gmail.com \\
       \addr Australian Artificial Intelligence Institute, University of Technology Sydney \\Ultimo, NSW 2007, Australia\\
       CFAR, Agency for Science, Technology and Research, 138632, Singapore
       \AND
       \name Ivor W. Tsang \email ivor.tsang@gmail.com\\
       \addr CFAR, Agency for Science, Technology and Research, 138632, Singapore\\
       IHPC, Agency for Science, Technology and Research,  138632, Singapore \\
       School of Computer Science and Engineering, Nanyang Technological University, 639798, Singapore\\
       Australian Artificial Intelligence Institute, University of Technology Sydney\\ Ultimo, NSW 2007, Australia
       }
\maketitle

\begin{abstract}

In the past few years, pre-trained language models (PTLMs) have brought significant improvements to target sentiment classification (TSC). Existing PTLM-based models can be categorized into two groups: 1) fine-tuning-based models that adopt PTLM as the context encoder; 2) prompting-based models that transfer the classification task to the text/word generation task. Despite the improvements achieved by these models, we argue that they have their respective limitations. For fine-tuning-based models, they cannot make the best use of the PTLMs’ strong language modeling ability because the pre-train task and downstream fine-tuning task are not consistent. For prompting-based models, although they can sufficiently leverage the language modeling ability, it is hard to explicitly model the target-context interactions, which are widely realized as a crucial point of this task.
In this paper, we present a new perspective of leveraging PTLM for TSC: simultaneously leveraging the merits of both language modeling and explicit target-context interactions via \textit{contextual target attributes}.
Specifically, we design the domain- and target-constrained cloze test, which can leverage the PTLMs' strong language modeling ability to generate the given target's attributes pertaining to the review context.
The attributes contain the background and property information of the target, which can help to enrich the semantics of the review context and the target.
To exploit the attributes for tackling TSC, we first construct a heterogeneous information graph by treating the attributes as nodes and combining them with (1) the syntax graph automatically produced by the off-the-shelf dependency parser and (2) the semantics graph of the review context, which is derived from the self-attention mechanism. Then we propose a heterogeneous information gated graph convolutional network to model the interactions among the attribute information, the syntactic information, and the contextual information.
The experimental results on three benchmark datasets demonstrate the superiority of our model, which achieves new state-of-the-art performance.
\end{abstract} 

\section{Introduction}
The task of target sentiment classification (TSC) \cite{TDLSTM,wwy_recursive} is to predict the sentiment of a given target included in a review.
In the review ``The noodles taste good, while the price is a little high.'', there are two targets (`noodles' and `price') with positive and negative sentiments, respectively.
Previous methods have realized that comprehensively understanding the review context and modeling the interactions between the context and the target are two crucial points for TSC.
And much effort has been paid to enhance them, including leveraging syntactic information \cite{asgcn,RGAT,dualgcn} and designing advanced attention mechanisms \cite{ATAE,AA-LSTM,Tencent}.
Besides, in recent years, since pre-trained language models (PTLMs) like BERT \cite{bert} have brought stunning improvement in a bunch of NLP tasks, current TSC models widely adopt the pre-trained BERT to encode the review context and then obtain the high-quality context hidden states, which have been proven to boost the performance significantly.
Most existing state-of-the-art (SOTA) TSC models are built on the BERT+Syntax+Attention framework.
More recently, prompt learning has attracted increasing attention in NLP fields.
It can transfer the classification task to the masked language modeling task.
In this way, the best potential of PTLMs can be utilized because the pre-training task and the downstream task are consistent (both are masked language modeling tasks) \cite{promptsurvey}.
And prompting-based methods have been explored in TSC
\cite{asprompt,openprompt}.

Therefore, existing PTLM-based models can be categorized into two groups: 1) fine-tuning-based models that adopt PTLM as the context encoder; 2) prompting-based models that transfer the classification task to the text/word generation task.
Despite the promising progress achieved in recent years, we argue that the two groups of models have their respective limitations. For fine-tuning-based models, it cannot make the best use of the PTLMs’ strong language modeling ability because the pre-train task and downstream fine-tuning task are not consistent. Although prompting-based models can sufficiently leverage the language modeling ability, it is hard to explicitly model the target-context interactions, which are widely realized as the key point of this task. 

In this paper, we simultaneously leverage the merits of masked language modeling (MLM) and explicit target-context interactions.
Unlike prompting-based methods leveraging MLM to generate the masked label word in the designed template, we mask the target in the review and obtain the generated words.
And in this paper, we refer to this process as the target cloze test.
After pre-training, the PTLMs are quite good language models, which can precisely generate the masked word fitting the semantics of the context well.
Given this fact, we believe the generated words contain the target's attribute information, which can help comprehensively understand the review context.

To investigate this, we conduct the target cloze test on some samples, and we list two of them in Table \ref{table: targe cloze}.
We can find that in the first sample, the generated words are generally related to the target \textit{Windows 7} and convey its property and background information.
However, in the second sample, the generated words seem random and irrelevant to the target \textit{browser}.
We attribute this to the fact that in the first sample, there is semantic constrain from \textit{Macbook pro laptop} and \textit{vmware program}, while there is none in the second sample.
Inspired by this discovery, we design the \textit{domain- and target-constrained cloze test}, which aims to generate the target-related candidate words that potentially provide the target's property and background information.
In this paper, we refer to these generated words as \textit{contextual target attributes} since they are generated regarding the review context and contain the target's attribute information.
We believe these contextual target attributes can benefit TSC via enhancing the context and target understanding.

\begin{table}[t]
\centering
\fontsize{10}{12}\selectfont
\setlength{\tabcolsep}{2mm}{
\begin{tabular}{l}
\toprule
 \begin{tabular}[c]{@{}l@{}}1. \textbf{\textit{Review}}: I also enjoy the fact that my Macbook pro laptop allows me to \\ run  [\textbf{Windows 7}]$^\text{masked}$
  on it by  using the vmware program. \end{tabular} \\ 
\begin{tabular}[c]{@{}l@{}} \textbf{\textit{Generated Words}}: games, linux, software, programs, applications, directly, ... \end{tabular} \\\midrule
 \begin{tabular}[c]{@{}l@{}}2. \textbf{\textit{Review}}: Speaking of the [\textbf{browser}]$^\text{masked}$, it too  has problems. \end{tabular} \\ 
 \begin{tabular}[c]{@{}l@{}} \textbf{\textit{Generated Words}}: past, sea, moon, weather, future, dead ... \end{tabular} \\
\bottomrule
\end{tabular}}
\caption{Two examples of the target cloze test, which is the vanilla cloze test masking the target. And the frozen BERT$_\text{base}$
is used to generate the candidate words.}
\label{table: targe cloze}
\end{table}

To exploit contextual target attributes for TSC, we propose a new model whose core is to model the interactions of three kinds of information: (1) the target's property and background information conveyed in the contextual target attributes; (2) the review's syntactic information conveyed by the syntax graph; (3) the review's contextual word correlative information conveyed by the fully-connected semantic graph derived from the self-attention mechanism.
Specifically, we first propose a heterogeneous information graph (HIG), which includes two kinds of nodes: attributes nodes and context nodes.
The connections among the context nodes are based on both the syntax graph and the semantic graph.
Moreover, the attribute nodes not only connect to the target node(s) but also connect to some context words via syntax-heuristic edges.
HIG provides a platform for attribute information-enhanced target-context interactions.
On HIG the fine-trained and heterogeneous interactions among the target's attribute information, target semantics and context semantics can be modeled.
To model the heterogeneous information interactions on HIG, inspired from \cite{darer,co-guiding,relanet,darer-pami,co-evolving,co-guiding-pami}, we propose a novel heterogeneous information gated graph convolutional network (HIG$^2$CN).
It includes a target- and context-centric gate mechanism to control the information flow.
Besides, relative position weights are adopted to highlight the potential target-related words.
In addition, we propose a heterogeneous convolution layer for information aggregation.
Generally, HIG$^2$CN has three advantages: 
\begin{itemize}
\item It leverages the contextual target attributes to enhance the context and target understanding, capturing more beneficial clues via the heterogeneous information interactions.
\item It integrates the structures of both syntax graph and semantic graph. Thus, it can capture more comprehensive and robust context knowledge.
\item Due to its gate mechanism, more and more crucial information can be discovered and retained in the nodes and then aggregated into the target node(s).
\end{itemize}
In this way, our method can simultaneously leverage the MLM via the target cloze test and explicitly model the target-context interactions via HIG$^2$CN.
We conduct evaluation experiments on three benchmark datasets. The results show that our model outperforms existing models by a large margin, achieving new state-of-the-art performance.

\section{Related Works} \label{sec: related work}
Different attention mechanisms \cite{ATAE,IAN,Tencent,songle2019,AA-LSTM} have been designed to capture the target-related semantics.
For example, IAN \cite{IAN} utilizes the interactive attention mechanism to model the bidirectional interactions between the target and context.
RAM \cite{Tencent} adopts a recurrent attention mechanism that calculates the attention weights recurrently to generally aggregate the target-related semantics.

Although attention mechanisms have achieved great improvements in TSC, in many cases, they still cannot capture the target-related semantics contained in some context words.
Recently, researchers discovered that the distance between the target and its related words can be shortened by applying graph neural networks (GNNs) to encode the syntax graph of the review context, which is a promising way to facilitate capturing the beneficial clues.
Therefore, different GNN-based models \cite{asgcn,DGEDT,sagat,tgcn,dualgcn,jair,RGAT,kagrmn,dignet} are proposed to model the syntax graph capturing the useful syntactic information that conveys the dependencies between the target and the crucial words.
For instance, ASGCN \cite{asgcn} applies a multi-layer graph convolutional network (GCN) on the syntax graph, which is obtained from an off-the-shelf dependency parser, to extract the syntactic information.
DualGCN \cite{dualgcn} is based on a dual-GCN architecture that captures both the semantic and syntactic information comprehensively.
DotGCN \cite{acl2022tree} integrates the attention scores and syntactic distance into a discrete latent opinion tree.

However, previous works leverage PTLM only for encoding. Our work makes the first attempt to exploit the contextual target attributes from PTLM. Besides, we propose the HIG and HIG$^2$CN to model the sufficient interactions among (1) the target's background and property information contained in the attributes, (2) the semantics information and (3) syntactic information.



\section{Overview}
\begin{figure}[t]
 \centering
 \includegraphics[width = 0.8\textwidth]{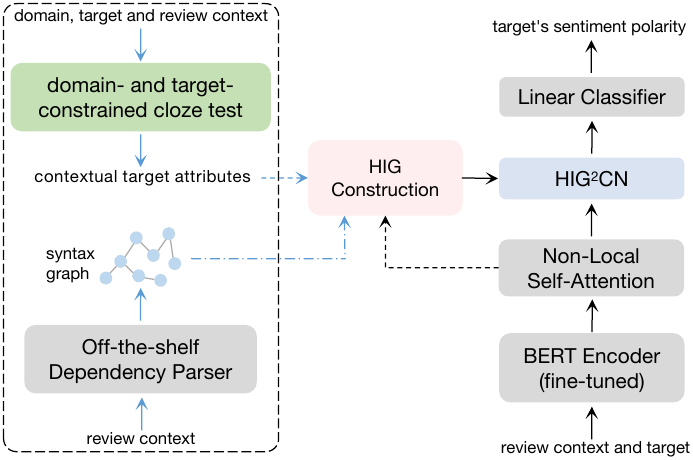}
 \caption{Overall architecture of our model. The dashed box denotes that the operations are conducted in the preprocessing procedure. The black dash arrow denotes the fully-connected semantic graph derived from the self-attention mechanism. HIG denotes heterogeneous information graph, and HIG$^2$CN denotes heterogeneous information gated graph convolutional network.}
 \label{fig: model}
\end{figure}

The overall framework of our method is illustrated in Fig. \ref{fig: model}.
In the preprocessing process, our proposed domain- and target-constrained cloze test is achieved by frozen BERT$_\text{base}$ and conducted on all samples to obtain the contextual target attributes, which are used to construct HIG.
Like previous methods, we utilize a off-the-shelf dependency parser\footnote{https://spacy.io/} to produce the syntax graphs of the samples.
In the training/testing process, after the BERT Encoder, the self-attention produces the attention matrix, which is adopted as the adjacent matrix of the fully-connected semantic graph.
Then the attributes, syntax graph, and semantic graph are fed to the HIG Construction module to generate the HIG, on which HIG$^2$CN is applied for heterogeneous information interactions.

Next, before diving into our model, we first introduce our proposed domain- and target-constrained cloze test.

\section{Domain- and Target-Constrained Cloze Test}

\begin{figure}[t]
 \centering
 \includegraphics[width = 0.7\textwidth]{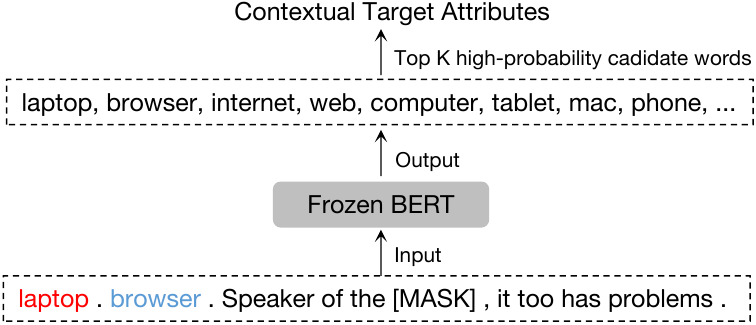}
 \caption{An example illustrating how to obtain the contextual target attributes. The red color denotes the domain word, and the blue color denotes the target. $K$ denotes the number of obtained contextual target attributes.}
 \label{fig: attribute}
\end{figure}
In this paper, we propose the domain- and target-constrained cloze test to obtain the contextual target attributes.
An example illustrating this process is shown in Fig. \ref{fig: attribute}.
To apply the semantics constraint from the domain and target to the input sentence, we design a simple while effective template in which the domain word and target are concatenated to the beginning of the masked review context.
From the second sample in Table \ref{table: targe cloze}, we can observe that the general cloze test conducted on BERT generates random and irrelevant words that cannot be used as attributes.
Differently, in Fig. \ref{fig: attribute}, we can find that the domain- and target-constrained cloze test can generate the target-related words (e.g., laptop, internet, web), which we term as \textit{contextual target attributes} conveying the background and property information of the target. 
Besides, they have another advantage that they naturally well-fit the context semantics since they are generated from the review context by the powerful PTLM.
Therefore, they have the potential to enhance target and context understanding.
Besides, the domain- and target-constrained cloze test is conducted on the frozen BERT, so this process can be conducted in the preprocessing process without costing extra training time.

\section{Our Model}

\subsection{Encoding}
BERT \cite{bert} is a popular pre-trained language model based on the multi-layer Transformer architecture. In this paper, BERT is employed for context encoding to obtain the context hidden states.
Denote the review context as $\{x_1, x_2, ..., x_{N_c}\}$ and the target words as $\{t_1, ..., t_{N_t}\}$, the input of BERT is the concatenation of these two sequences:
\begin{equation}
  \langle[\text{CLS}]; {x_1, x_2, ..., x_{N_c}}; \text{[SEP]}; {t_1, ..., t_{N_t}}; [\text{SEP}]\rangle
\end{equation}
 where [CLS]\footnote{CLS denotes classification. The [CLS] token is added at the beginning of the input sequence and its representation is usually taken as the sentence-level representation for classification.} and [SEP] are the special tokens in BERT; $\langle;\rangle$ denotes sequence concatenation;
 $N_t$ and $N_c$ denote the word number of the target and review context, respectively.
We take the last-layer representation in BERT as the context word hidden states: $\hat{H_c}=[\hat{h}^c_1, ..., \hat{h}^c_{N_c}]\in\mathbb{R}^{N_c \times d}$, which includes the target hidden states $\hat{H_t} = [\hat{h}^t_1, ..., \hat{h}^t_{N_t}]\in\mathbb{R}^{N_t \times d}$, where $d$ denotes the hidden state dimension.
Besides, due to BERT's sentence-pair modeling ability, it has been proven that the [CLS] token's output hidden state $h_{cls}$ can capture the target-context dependencies \cite{RGAT,zhangyueRGAT,jair}.
Therefore, we adopt $h_{cls}$ as the target-centric context semantics representation, which is used in Eq. \ref{eq: context gating}.

\subsubsection{Non-Local Self-Attention}
Although retrieving BERT's self-attention matrix at the last layer seems an alternative way to obtain the self-attention graph, this approach cannot be used to construct the fully-connected semantic graph.
The reason is that the Transformer blocks in BERT include many \textit{multi-head} self-attention mechanisms, which segment word representations into multiple \textit{local} subspaces.
Therefore, the self-attention matrix corresponding to each head can only represent the \textit{biased} word correlations in the \textit{local} subspace.
However, all of the subsequent modules work in the \textit{global} semantic space.

To solve this, we employ a non-local self-attention layer after the BERT encoder so as to generate the \textit{global} self-attention matrix $A^{sat}\in\mathbb{R}^{N_c \times N_c}$, which is the adjacent matrix of the semantic graph and represents the contextual word correlative information.
Specifically, $A^{sat}$ is obtained by:
\begin{equation}
  {A^{sat}} = \operatorname{Softmax}\left(\frac{(\hat{H_c} M_q) ({\hat{H_c} M_k })^{\mathsf{T}}}{\sqrt{d}}\right) 
\end{equation}
The information aggregation is conducted along $A^{sat}$ to update the hidden states:
\begin{equation}
  H=A^{sat} \hat{H_c} M_v
\end{equation}
where $M_* \in \mathbb{R}^{d\times d}$ are parameter matrices.

The target representation is obtained by applying average pooling over the target hidden states:
\begin{equation}
  r^t=\frac{1}{N_t}\sum_i^{N_t}{h^t_i} \label{eq: target rep}
\end{equation}
\subsection{Heterogeneous Information Graph Construction}
\begin{figure}[t]
 \centering
 \includegraphics[width = 0.75\textwidth]{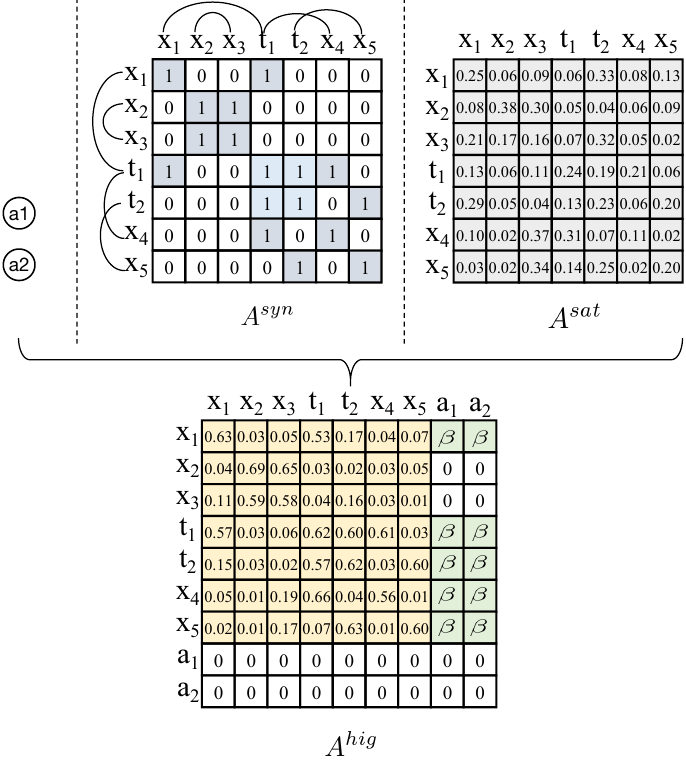}
 \caption{Illustration of the process of HIG construction. In this example, w.l.o.g, there are two contextual target attributes: $a_1$ and $a_2$; the coefficient $\alpha$ balancing the syntax graph and semantics graph is 0.5. And the arrows denote the edges/dependencies between the words on the syntax graph. }
 \label{fig: hig construct}
\end{figure}
In this paper, we design the heterogeneous information graph (HIG) to integrate the background and property information of context target attributes, the syntactic information conveyed by the syntax graph, as well as the contextual word correlative information conveyed by the semantics graph.
Fig. \ref{fig: hig construct} illustrates the process of HIG construction.
In the preprocessing stage, 
the syntax graph is automatically produced by the off-the-shelf dependency parser.
Then we regard the syntax graph as an undirected graph, which is consistent with previous works \cite{asgcn,jair}, and an example of its adjacent matrix $A^{syn}$ is shown in Fig. \ref{fig: hig construct}.
Note that in $A^{syn}$, we add the edges between the target words to let them fully connect to each other, aiming to enhance the target understanding.

In the training/testing process, the HIG is constructed based on the contextual target attributes, $A^{syn}$ and $A^{sat}$.
We first merge $A^{syn}$ and $A^{sat}$ with a coefficient $\alpha$ balancing the syntactic information and the contextual word correlative information:
\begin{equation}
  \hat{A}^{hig}_{i,j} = \alpha \cdot A^{syn}_{i,j} + (1-\alpha) \cdot A^{sat}_{i,j}
\end{equation}

Then we add the nodes of contextual target attributes regarding the following rules:\\
(1) There are edges between the attribute nodes and the target word nodes.
This can enhance the target representation via integrating the background and property information contained in the attributes.\\
(2) There are edges between the attribute nodes and the context word nodes whose corresponding words are connected to the target words in the syntax graph.
Intuitively, the context words that are syntactically related to the target, can usually help to infer the target's sentiment, and this is the reason that most recent models leverage the syntax graph.
Therefore, integrating the background and property information contained in the attributes into the syntactically related context words can help to understand the context semantics more comprehensively, discovering more beneficial clues.
\\
(3) The above edges are unidirectional and directed from the attribute nodes to the target nodes or the context word nodes.
The intuition behind this is that the information contained in the attribute nodes should be reserved, and they are supposed to provide the `pure' information to other nodes.
Therefore, HIG does not include the edges directed into the attribute nodes to guarantee that in the information aggregation process, they do not receive information from other nodes.\\
(4) There is a coefficient $\beta$ to control the weight of the edges related to the attribute nodes.
The edge weights between the context words are between 0 and 1.
Intuitively, too large weights of the attribute nodes' edges would dilute the syntactic information and contextual word correlative information, while too small weights could not integrate enough background and property information of the target.
Therefore, we use a hyper-parameter $\beta$ to control the weight.

Then we can obtain the final HIG and its adjacent matrix $A^{hig}$. An example of $A^{hig}$ is shown in Fig. \ref{fig: hig construct}.

\subsection{Heterogeneous Information Gated Graph Convolutional Network}
\begin{figure}[t]
 \centering
 \includegraphics[width = 0.9\textwidth]{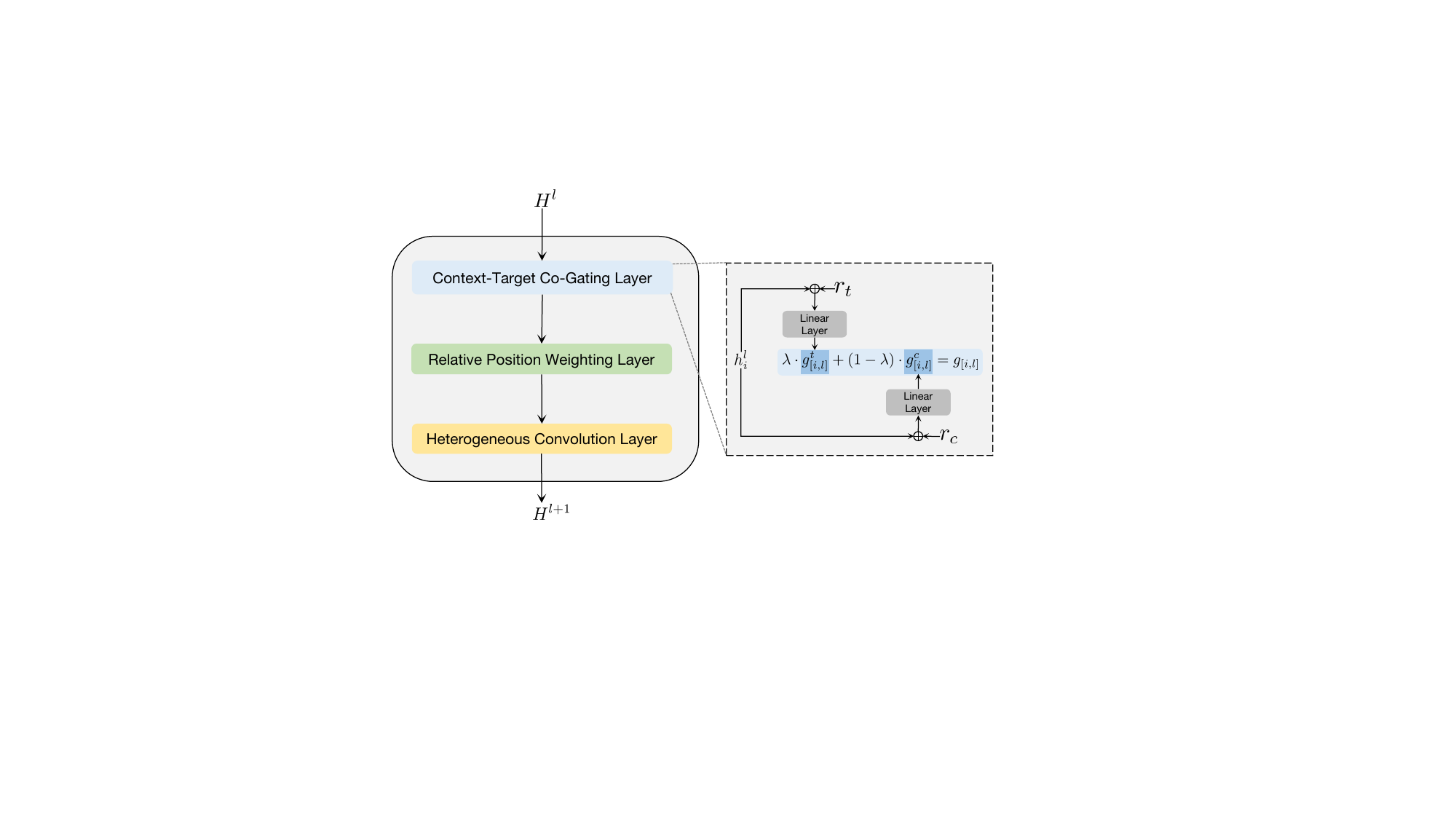}
 \caption{Illustration of a single HIG$^2$CN layer.}
 \label{fig: hig2cn}
\end{figure}
Fig. \ref{fig: hig2cn} illustrates the architecture of a single HIG$^2$CN layer, which includes three sub-layers. 

\subsubsection{Context-Target Co-Gating Layer}
We design a context-target co-gating layer to adaptively control the information flow regarding the target semantics and the target-centric context semantics.
The objective is to distill beneficial information and filter out noisy information in the information aggregation process of HIG$^2$CN.
The inner details of the context-target co-gating layer are shown in Fig. \ref{fig: hig2cn}.
Specifically, it can be formulated as:
\begin{equation}
\begin{split}
g^t_{[i,l]} &= W_g^t(h_i^l\oplus r_t) + b_g^t\\
g^c_{[i,l]} &= W_g^c(h_i^l\oplus r^t_c) + b_g^c\\ \label{eq: context gating}
g_{[i,l]} &=\sigma\left(\lambda \cdot g^t_{[i,l]} + (1-\lambda) \cdot g^c_{[i,l]}\right)
\end{split}
\end{equation}
where $g^t_{[i,l]}$, $g^c_{[i,l]}$ and $g_{[i,l]}$ are the target gating vector, context gating vector and the final co-gating vector, respectively; $W_g^*$ and $b_g^*$ denote the weight and bias, respectively; $\oplus$ denotes vector concatenation operation; $\sigma$ denotes the sigmoid function; $\lambda$ denotes the hyper-parameter balancing the impacts of the target and context to the co-gating vector $g_{[i,l]}$; $r_t$ denotes the target semantics, which is obtained in Eq. \ref{eq: target rep}; $r^t_c$ denotes the target-centric context semantics, and in this paper, it is $h_{cls}$, which has been obtained after encoding.

\subsubsection{Relative Position Weighting Layer}
Intuitively, the target-related words are usually close to the target words in the review context.
Therefore, we design a relative position weighting layer, which calculates the weight of each context word regarding its relative distance to the target words, aiming to highlight potential target-related words.
Specifically, the weight is calculated by:
\begin{align}
    w_p^j =& 1 - \frac{\min \limits_{\mu\leq\tau\leq \mu+N_t } \vert j-\tau \vert}{N_c+1} \label{eq: pos weight}
\end{align}
where $\mu$ denotes the first target word's position in the context.
\subsubsection{Heterogeneous Convolution Layer}
In previous models, GCN \cite{gcn} is widely adopted for syntax graph modeling,
which can be formulated as:
\begin{equation}
h_i^{l+1}=\operatorname{ReLU}\left(\sum_{j=1}^{N_c} {A}_{i j} {W}_1^{l} {h}_{j}^{l-1}/\left(d_{i}+1\right)+{b}^{l}\right)
\end{equation} 
where $A$ denotes the adjacent matrix; $W^l_1$ and $b^l$ denote the weight and bias; $d_i$ denotes the degree of node $i$.

However, it is inappropriate to adopt vanilla GCN to achieve the information aggregation on our HIG.
On the one hand, it cannot take the co-gating vector $g_{[i,l]}$ and the relative position weight into consideration.
On the other hand, it is intuitive that when a context node receives information, an attribute node's information and another context node's information have different contributions, while GCN cannot handle this.
Therefore, in this paper, we propose the heterogeneous convolution layer to achieve the information aggregation in our HIG, and it can be formulated as:
\begin{equation}
\begin{split}
h_i^{l+1}&=\operatorname{ReLU}\left(\sum_{j=1}^{N_c} A^{hig}_{i j} W_1^{l} \hat{h_j^l} /\left(d_{i}+1\right)+b^{l}\right)\\
\hat{h_j^l} &= g_{[j,l]} \odot(W^{\phi(j)}h_{j}^l) \cdot w^{\phi(i)}\\
{W}^{\phi(j)} &= 
\begin{cases}
{W}^a & {\phi(j) = attribute}\\
{I} & {\phi(j) \neq attribute}
\end{cases} \\
w^{\phi(j)} & = 
\begin{cases}
1 & {\phi(j) = attribute}\\
w_p^i & {\phi(j) \neq attribute}
\end{cases} \\
\end{split}
\end{equation} 
 $A^{hig}$ denotes the adjacent matrix; $W^l_1$ and $b^l$ denote the weight and bias; $d_i$ denotes the degree of node $i$.
$\odot$ denotes the Hadamard product operation.
$\phi(j)$ is a function that identifies whether node $j$ is an attribute node.
${W}^{\phi(j)}$ is the transformation matrix that discriminates the attribute nodes and context nodes by projecting their representations into different spaces.
For simplification, if node $j$ is an attribute, we use a trainable weight matrix $W^a$ to project its representation; if node $j$ is not an attribute node, it is multiplied by the identity matrix $I$.
$w^{\phi(j)}$ is the relative position weight of node $j$.
If node $j$ is an attribute node, it does not have the relative position weight because it is not included in the review context, so we set it to 1.
Actually, the weight of the attribute node is controlled by the hyper-parameter $\beta$.
If node $j$ is not an attribute node, it has the relative position weight $w_p^j$ that is calculated by Eq. \ref{eq: pos weight}. 

By this means, HIG$^2$CN can model the interactions among the background and property information of the target, the syntactic information and the contextual word correlative information.
And in the information aggregation process, the information flow is adaptively controlled by the target's semantics and the target-centric context's semantics.
After the stacked $L$ layers of HIG$^2$CN, the target word nodes are supposed to contain sufficient target sentiment clues.
Then we obtain the final target representation by applying average pooling over all target word nodes:
\begin{equation}
  R=\frac{1}{N_t}\sum_i^{N_t}\!{h^t_{[i,L]}} \label{eq: final target rep}
\end{equation}

\subsection{Prediction and Training Objective}
We first concatenate $R$ with $h_{cls}$, and then the softmax classifier is used to produce the sentiment label distribution:
\begin{equation}
P = \operatorname{Softmax}(\text{MLP}(R\oplus h_{cls}))
\end{equation}
where MLP denotes the multi-layer perception layer.

Then we apply the argmax function to obtain the final predicted sentiment label:
\begin{equation}
  \hat{y} = \operatorname{argmax} \limits_{k\in\mathcal{S}}P[k]
\end{equation}
where $\mathcal{S}$ denotes the set of sentiment classes.

For model training, the standard cross-entropy loss is employed as the objective function:
\begin{equation}
  \mathcal{L} = -\sum_{i=1}^{D}\operatorname{log}(P[\mathcal{C}_i])
\end{equation}
where $D$ denotes the number of training samples; $\mathcal{C}_i$ denotes the index of the $i$-th training sample's golden label.

\section{Experiment}\label{sec:experiment}
\subsection{Benchmarks and Implementation Details}
We conduct experiments on the Restaurant14, Laptop14 and Restaurant15 datasets \cite{Semeval2014,semeval2015}, which are widely adopted test beds for the TSC task.
We pre-process the datasets following previous works \cite{asgcn,RGAT,tgcn}.
Table \ref{table: dataset} lists the detailed statistics of the datasets.

\begin{table}[ht]
\centering
\fontsize{10}{12}\selectfont
\setlength{\tabcolsep}{1.5mm}{
\begin{tabular}{ccccccc}
\toprule
\multirow{2}{*}{Dataset} & \multicolumn{2}{c}{Positive} & \multicolumn{2}{c}{Neutral} & \multicolumn{2}{c}{Negative} \\\specialrule{0em}{0pt}{1pt} \cline{2-3} \cline{4-5} \cline{6-7}\specialrule{0em}{1pt}{1.5pt}
                         & Train         & Test         & Train         & Test        & Train         & Test         \\ \specialrule{0em}{0pt}{1pt}\hline \specialrule{0em}{1.5pt}{1.5pt}
Laptop14                    & 994           & 341          &  464          &    169      &  870          &    128       \\
Restaurant14                    & 2164          & 728          & 637           & 196         & 807           & 196          \\
Restaurant15                    & 912           & 326          & 36            & 34          & 256           & 182         \\\bottomrule
\end{tabular}}
\caption{Dataset statistics.}
\label{table: dataset}
\end{table}

We adopt the BERT$_{\text{base}}$ uncased version\footnote{Layer number is 12; hidden dimension is 768; attention head number is 12; total parameter number is 110M} for both encoding and the domain- and target-constrained cloze test. The BERT encoder is fine-tuned in the training process\footnote{The final loss of our model is back-propagated to not only HIG$^2$CN's parameters but also the loaded BERT's parameters. In this way, BERT is fine-tuned to generate better contextual hidden states.}.
We adopt the AdamW optimizer \cite{weightdecay}, which is used to train our model.
We use the off-the-shelf dependency parser from the spaCy toolkit to obtain the syntax graph of the review context.
Table \ref{table: hyper-parameter} lists the details of the hyper-parameters. 

\begin{table}[ht]
\centering
\fontsize{10}{12}\selectfont
\setlength{\tabcolsep}{1.5mm}{
\begin{tabular}{l|l}
\toprule
Learning Rate & 1e-5 \\  \hline
Dropout Rate & 0.3 \\ \hline
Weight Decay Coefficient & 0.05 \\ \hline
Layer number of  HIG$^2$CN & 3 \\
Dimension of Hidden State & 768 \\
Coefficient $\alpha$ & 0.5 \\
Coefficient $\beta$ & 0.5
\\\bottomrule
\end{tabular}}
\caption{Dataset statistics.}
\label{table: hyper-parameter}
\end{table}

Accuracy and Macro-F1 are used as evaluation metrics.
Following previous works \cite{asgcn,DGEDT,zhangyueRGAT}, we report the average results over three runs with random initialization.

\subsection{Baselines for Comparison}
We compare our model with the following five groups of baselines. The models in the first four groups are fine-tuning-based and the ones in the last group are prompting-based.

(A) Neither BERT nor Syntax is leveraged:\\
1. IAN \cite{IAN}. It leverages the proposed interactive attention mechanism, which calculates the target words' attention weights based on context semantics and the context words' attention weights based on the target semantics. \\
2. RAM \cite{Tencent}. It employs the recurrent attention mechanism that recurrently calculates the attention weights and aggregates the target-based semantics to the cell state.

(B) BERT is not leveraged while Syntax is leveraged:\\
3. ASGCN \cite{asgcn}. It applies multi-layer GCN over a syntax graph to capture syntactic information.\\
4. BiGCN \cite{bigcn}. It applies the GCN over the constructed hierarchical syntactic and lexical graphs.

(C) BERT is leveraged while Syntax is not leveraged:\\
5. BERT-SPC \cite{bert}. Taking the concatenated context-target pair as input, this model uses the output hidden state of the [CLS] token for classification.\\
6. AEN-BERT \cite{aen-bert}. It stacks the attention layers in a multi-layer manner to learn deep target-context interactions.

(D) Both BERT and Syntax are leveraged:\\
7. ASGCN+BERT \cite{asgcn}. Based on ASGCN, we replace the original LSTM encoder with the same BERT encoder used in our model.\\
8. KGCapsAN-BERT \cite{kgcap}. It leverages different kinds of knowledge to enhance the capsule attention, and the syntactic knowledge is obtained by applying GCN over the syntax graph. \\
9. R-GAT+BERT \cite{RGAT}. It captures the global relational information by considering both the target-context correlation and syntax relations. \\
10. DGEDT-BERT \cite{DGEDT}. It models the interactions between flat textual semantics and syntactic information through the proposed dual-transformer network.  \\
11. RGAT-BERT \cite{zhangyueRGAT}. It leverages the dependency types between the context words to capture comprehensive syntactic information.\\
12. A-KVMN+BERT \cite{kvmn-eacl}. It leverages both word-word correlations and their syntax dependencies through the proposed key-value memory network. \\
13. BERT+T-GCN \cite{tgcn}. It models the dependency types among the context words by the proposed T-GCN and obtains a comprehensive representation via layer-level attention and attentive layer ensembling. \\
14. DualGCN+BERT \cite{dualgcn}. It models the interactions between semantics and syntax by the proposed orthogonal and differential regularizers. \\
15. RAM+AAGCN+AABERT3 \cite{jair}; This model augments RAM with the aspect-aware BERT and aspect-aware GCN to enhance the aggregation of target-related semantics.\\
16. dotGCN \cite{acl2022tree}. This model employs a discrete latent opinion tree to augment the explicit dependency trees.

(E) {Prompting-based Models}: \\
17. BERT NSP Original and BERT LM Original \cite{openprompt}. These two models explore language modeling (LM) prompts and natural language inference (NLI) prompts, respectively.\\
18. AS-Prompt \cite{asprompt}. This model adopts continuous prompts to transfer the sentiment classification task into Masked Language Modeling (MLM) by designing appropriate prompts and searching for the ideal expression of prompts in continuous space.

In all of the above baselines, the BERT encoders are the same as ours, adopting the BERT-base uncased version.
For fair comparisons, following Bai et al., 2021, we do not include the baselines that exploit external resources such as auxiliary sentences \cite{auxiliarybert}, extra corpus \cite{bert-post-train} and knowledge base \cite{kagrmn,arbert}.
Besides, since R-GAT+BERT, RGAT-BERT, A-KVMN+BERT, BERT+T-GCN and DualGCN+BERT did not report the average results in their original papers, we reproduce their average results on three random runs.

\subsection{Main Result}
\begin{table*}[t]
\fontsize{10}{12}\selectfont
\centering
\caption{Comparison of evaluation results on the three benchmark datasets (in \%). $^\dag$ indicates the results are reproduced using the official source code. $^\ddag$ denotes that the results were reported by Zhang et al., 2019. The best scores are in \textbf{bold} and the second best scores are \underline{underlined}.
Our model significantly outperforms baselines on all datasets and all metrics ($p < 0.05$ under t-test).
For the t-test of our model and dotGCN on average F1, the calculated t-statistic is 5.57 and the degrees of freedom are 4.}
\setlength{\tabcolsep}{1.8mm}{
\begin{tabular}{c|cccccc|cc}\toprule
 \multirow{2}{*}{Models} & \multicolumn{2}{c}{Laptop14} & \multicolumn{2}{c}{Restaurant14} & \multicolumn{2}{c|}{Restaurant15} &\multicolumn{2}{c}{Average}  \\\cline{2-9} 
      &Acc      &F1       &Acc       &F1        &Acc         &F1     &Acc         &F1    \\\midrule
IAN$^\ddag$ &72.05     & 67.38    &79.26    &70.09  & 78.54    & 52.65 &76.62 &63.37 \\
RAM$^\ddag$ &74.49& 71.35 & 80.23 &70.80&79.30&60.49 &78.01 & 67.55\\ \hline
 ASGCN & 75.55    &71.05     &80.77    & 72.02  & 79.89   & 61.89  &78.74 &68.32  \\
 BiGCN & 74.59    &71.84    &81.97     &73.48    & 81.16    & 64.79  &79.24 &70.04  \\ \hline
 BERT-SPC& 78.47    &73.67  &84.94   & 78.00   & 83.40    &  65.00   &82.27 & 72.22 \\
 AEN-BERT  & 79.93    &76.31  &83.12   & 73.76   & -    &  -  &- &-   \\ \hline
 ASGCN+BERT$^\dag$    & 78.92    &74.35     &85.87    & 79.32  & 83.85   & 68.73 &82.88 &   74.13      \\
KGCapsAN-BERT  & 79.47    &76.61    &85.36      & 79.00   & - & - &- &-  \\
R-GAT+BERT$^\dag$& 79.31    &75.40  &86.10   & 80.04   & 83.95    &  69.47  &83.12 &74.97   \\
DGEDT-BERT  & 79.8    &75.6    &86.3      & 80.0   & 84.0   &71.0   &83.37 &75.53   \\
 A-KVMN+BERT$^\dag$  & 79.20    &75.76   &85.89      & 78.29   & 83.89  &67.88  &82.99 &73.98    \\
 BERT+T-GCN$^\dag$  & 80.56    &76.95    &85.95      & 79.40   & 84.81 & 71.09 &83.65 & 75.81 \\
 DualGCN+BERT$^\dag$  & 80.83 & 77.35   & 86.64  &  80.76 &84.69&71.58 & 84.05& 76.56 \\
 RGAT-BERT &80.94 & \underline{78.20} & 86.68 & \underline{80.92} &- &- &- &-\\
   RAM+AAGCN+AABERT3     &81.20 & 78.01 &\underline{86.79} & 80.69 & - & -  &- &-  \\
 dotGCN & 81.30    & 78.10    & 86.16     & 80.49   & \underline{85.24} & \underline{72.74} & \underline{84.23}&\underline{77.11}  \\ \hline
BERT NSP Original & \underline{84.77} & 76.93   &77.96& 73.24&-&-&-&-\\
BERT LM Original & 84.26 &75.90 &  77.61 & 73.06 &-&-&-&- \\
AS-Prompt &\textbf{85.64} & 76.71 & 78.37 & 74.19&-&-&-&- \\
\midrule
BERT+HIG$^2$CN (ours) & 82.18   &\textbf{78.70}     &\textbf{87.50}       & \textbf{82.05}    &\textbf{86.53} &\textbf{75.71} &\textbf{85.40} &\textbf{78.82}   \\
standard deviation & $\pm$0.35&$\pm$0.49&$\pm$0.28&$\pm$0.47&$\pm$0.52& $\pm$0.69& - &- \\
\bottomrule
\end{tabular}}

\label{table: results}
\end{table*}
The result comparison of our BERT+HIG$^2$CN and the five groups of baselines on the three datasets are shown in Table \ref{table: results}.
We can observe that BERT can significantly boost performance, and the basic model BERT-SPC obviously overpasses all LSTM-based models.
Nevertheless, although BERT can effectively capture contextual knowledge, augmenting BERT with syntactic information can bring further significant improvement.
Although our model is based on the same BERT+Syntax+Attention framework as the existing best models, it obtains consistent improvements over them on both Acc and F1.
Compared with the up-to-date best model dotGCN, in terms of Acc, our model gains improvements of 1.1\%, 1.6\%, and 1.5\%  on the Laptop14, Restaurant14, and Restaurant15 datasets, respectively.
Additionally, in terms of F1, our model gains improvements of 0.8\%, 1.9\%, and 4.1\% on the Laptop14, Restaurant14, and Restaurant15 datasets, respectively.
As for the average performance, our model surpasses dotGCN by 1.4\% and 2.2\% in terms of Acc and F1, respectively.
We can find that the prompting-based models usually obtain worse results compared with the BERT+Syntax+Attention models.
We suspect the main reason is that although prompting-based models can make the best use of the language modeling ability, they cannot explicitly model the target-context interactions, which is crucial for TSC.
Compared with prompting-based models, our model achieves consistent and significant improvements.
The promising results of our model can be attributed to the fact that we exploit the contextual target attributes from BERT via the domain- and target-constrained cloze test, and then effectively leverage them for comprehensive target and context understanding.
Note that all baselines utilize the attention modules to extract the target-related sentiment information, while there is no attention module in our model.
Therefore, the promising performance of our model proves that our HIG$^2$CN can effectively and adaptively capture the target-related sentiment information via modeling the heterogeneous information interactions among the target's background and property information in the attribute, the syntactic information, and the contextual word correlative information.

\subsection{Ablation Study}
\begin{table}[ht]
\centering
\fontsize{10}{12}\selectfont
\setlength{\tabcolsep}{2mm}{
\begin{tabular}{c|cccccc}
\toprule
\multirow{2}{*}{Variants} & \multicolumn{2}{c}{Laptop14} & \multicolumn{2}{c}{Restaurant14} & \multicolumn{2}{c}{Restaurant15}  \\\cline{2-7} 
      &Acc      &F1      &Acc       &F1       &Acc       &F1       \\\midrule
Full Model  &\textbf{82.18}   &\textbf{78.70}     &\textbf{87.50}       & \textbf{82.05}    &\textbf{86.53} &\textbf{75.71} \\ \midrule
{NoAttribute}      &81.05 &77.34    &86.48     &80.65     &85.06    &71.51 
                              \\ 
{NoSyntax}    &81.33  &77.69 &86.57   &80.56  &85.24   &71.74   
                               \\ 
{NoSemantics}    &81.70  &78.19 &86.85   &81.26  &85.67   &73.05   
                              \\ \midrule
{NoHIG$^2$CN}     &81.03 &77.44    &86.63     &80.81     &84.87    &71.57 
                              \\ 
{NoContextGating}    &81.59  &77.97 &86.93   &81.34  &85.98   &73.70   
                               \\ 
 {NoTargetGating}   &81.78  &77.84 &87.02   &81.16  &86.04   &73.82  
                              \\
\bottomrule
\end{tabular}}
\caption{Results of ablation experiments. Our full model significantly outperforms the ablated variants on all datasets and all metrics (p $<$ 0.05 under t-test).}
\label{table: ablat}
\end{table}
There are two cores of our model: HIG and HIG$^2$CN.
To verify the necessity of their modules, we conduct two groups of ablation experiments, whose results are shown in Table \ref{table: ablat}.

\paragraph{HIG} 
There are three sources of HIG: contextual target attributes, syntax graph and semantics graph.
To study their effects, we design three variants: \textit{NoAttribute}, \textit{NoSyntax} and \textit{NoSemantics}, respectively.
\textit{NoAttribute} is implemented by removing the attribute nodes on HIG.
\textit{NoSyntax} is implemented by directly adding the attribute nodes to the semantics graph.
\textit{NoSemantics} is implemented by directly adding the attribute nodes to the syntax graph.
From Table \ref{table: ablat}, it can be seen that without attributes, the performance drops significantly, indicating that integrating the 
contextual target attributes to leverage the target's background and property information can benefit the model.
Another observation is that leaving out the syntax graph also leads to worse performance, proving that leveraging syntactic information is indispensable for the model because it can effectively enhance the context understanding to capture the target-related information.
Moreover, we can find that if the semantics graph is not considered, the performance also drops sharply.
This is because the semantics graph includes the correlation scores between every two words, which can help to capture more comprehensive contextual information.

\paragraph{HIG$^2$CN}
To verify the effectiveness of HIG$^2$CN, we design a variant \textit{NoHIG$^2$CN} via replacing HIG$^2$CN with vanilla GCN.
From Table \ref{table: ablat}, we can observe that \textit{NoHIG$^2$CN} obtains dramatic performance degradation, proving that HIG$^2$CN is crucial for the whole model to capture the beneficial clues for TSC via heterogeneous information interactions.
This can be attributed to the fact that (1) the target-context co-gating layer in HIG$^2$CN can adaptively control the information flow regarding the target and the target-centric context semantics; (2) HIG$^2$CN can achieve more appropriate information aggregation via discriminating the information from attribute nodes and context nodes.
To investigate the target-context co-gating layer, we design two variants: \textit{NoContextGating} and \textit{NoTargetGating}, which are implemented by removing the context gating vector $g^c_{[i,l]}$ and the target gating vector $g^t_{[i,l]}$, respectively.
Compared with the full model, both \textit{NoContextGating} and \textit{NoTargetGating} obtain worse results.
This proves that both the target semantics and the target-centric context semantics can provide indicating information for extracting beneficial information and filtering out noisy information in the information aggregation process.

\subsection{Effect of Attribute Number}
\begin{figure}[t]
 \centering
 \includegraphics[width = 0.9\textwidth]{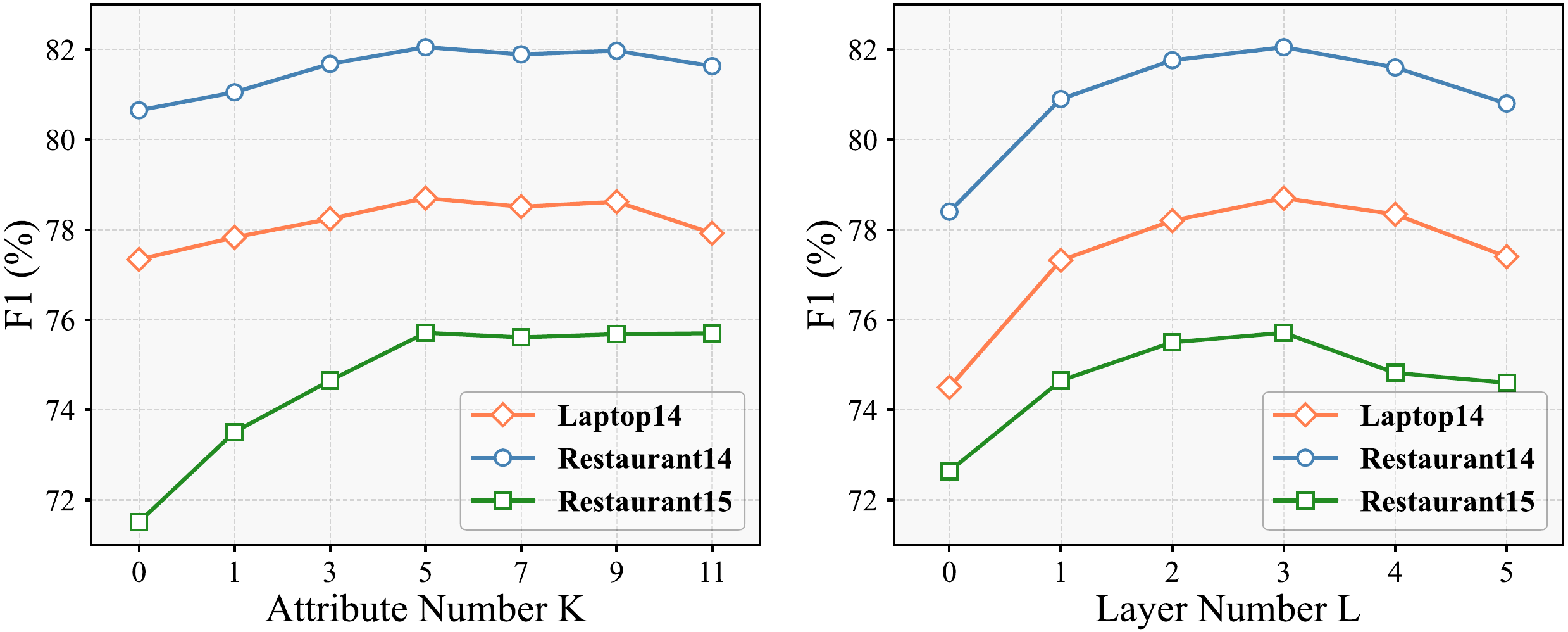}
 \caption{F1 score comparison with different contextual target attribute number ($K$) and HIG$^2$CN layer number ($L$).}
 \label{fig: num_range}
\end{figure}
To study the effect of attribute number, we vary its value in the set of \{0,1,3,5,7,9,11\}, and the results (in F1) are shown in Fig. \ref{fig: num_range}.
We can observe that the results first increase and then stop rising or even drop when $K>5$.
We suppose there are two reasons.
First, large $K$ results in some low-confidence attributes containing noisy information.
Besides, too many attributes lead to massive attribute information in HIG$^2$CN, and then dilute the beneficial syntactic information and the contextual word correlative information.

\subsection{Effect of HIG$^2$CN Layer Number}
We investigate the effect of HIG$^2$CN layer number by setting it with different values ranging from 0 to 5, and the results are shown in Fig. \ref{fig: num_range}.
On all datasets, our model achieves the best result when $L=3$.
And with $L$ increasing from 0 to 5, the performance first increases and then drops.
This indicates that although more layers of HIG$^2$CN can learn deeper interactions of the target's background and property information, the syntactic information, and the contextual word correlative information,
too large $L$ leads to inferior performance since it causes the over-smoothing and over-fitting problems.

\subsection{Case Study}

\begin{table*}[ht]
\centering
\fontsize{9}{11}\selectfont
\setlength{\tabcolsep}{1mm}{
\begin{tabular}{l|l|l|l}
\toprule
 Review & Context Target Attributes & Full Model & NoAttribute \\ \midrule
  \begin{tabular}[c]{@{}l@{}}1. \textbf{Boot time} is super fast , around anywhere \\ from 35 seconds to 1 minute .
 \end{tabular} 
&  \begin{tabular}[c]{@{}l@{}}boot, speed, startup, drive, \\ access \end{tabular}   & Positive ($\checkmark$) & Neutral ($\times$)\\ \midrule
 \begin{tabular}[c]{@{}l@{}}2. Entrees include classics like lasagna ,  \\ \textbf{fettuccine alfredo} and chicken parmigiana.
 \end{tabular} 
&  \begin{tabular}[c]{@{}l@{}}pasta, spaghetti, tomato,\\ pizza, salad \end{tabular}   & Neutral ($\checkmark$)  & Positive ($\times$)\\
\bottomrule
\end{tabular}}
\caption{Case study. The target words are in \textbf{bold}. The third and fourth columns denote the sentiment predicted by our full model and the NoAttribute variant, respectively. Sample 1 is from the Laptop14 test set (including 341 test samples in the positive class). Sample 2 is from the Restaurant14 test set (including 196 test samples in the neutral class).}
\label{table: case study}
\end{table*}

To understand how context target attributes benefit TSC, we present a case study with two test samples, as shown in Table \ref{table: case study}.
In the first sample, without attributes, \textit{NoAttribute} predicts the neutral sentiment regarding the semantics of `\textit{time is fast}'. Augmented with the attributes (e.g., speed, startup), our model can give the correct prediction regarding the semantics of `\textit{speed/startup is fast}', which is generally positive.
In the second sample, \textit{NoAttribute} may hardly understand the target `fettuccine alfredo', then predicts the positive sentiment via wrongly regarding `\textit{like}'.
Given the attributes, our model is aware that this sentence just introduces the different foods included in the entrees without expressing sentiment.
Then it can predict the correct sentiment.

\section{Discussion}

\begin{figure}[t]
 \centering
 \includegraphics[width = \textwidth]{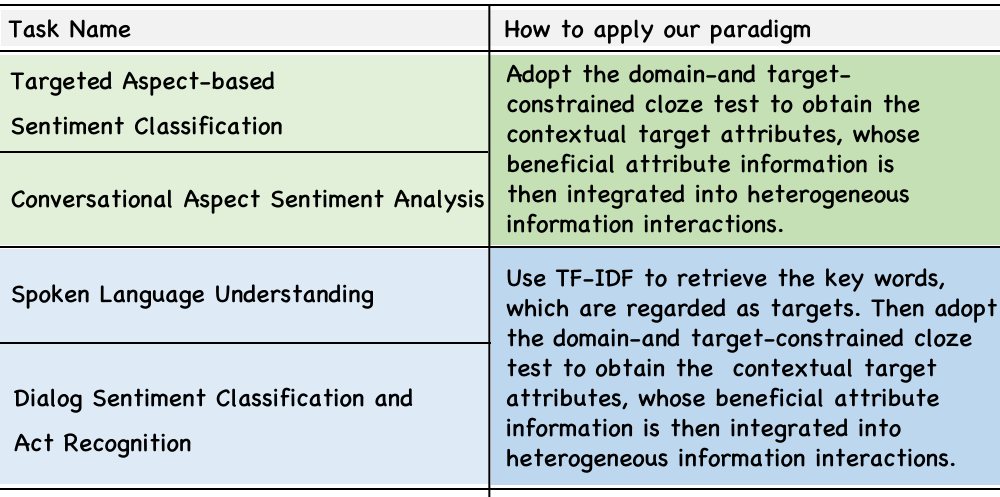}
 \caption{The first two tasks are representatives of target-based tasks, which have given targets.
 The second two tasks are representatives of general tasks that do not have given targets. 
 For these general tasks, we propose to use TF-IDF to retrieve the topic/key words, which are taken as targets. 
 And then our method can be applied.}
 \label{fig: generlize}
\end{figure}

Essentially, our approach is to retrieve the target-based contextual knowledge from the PTLM and then integrate this knowledge or target attribute information in the reasoning process. Our approach can be generalized into not only target-based tasks but also more general tasks.
In Fig. \ref{fig: generlize}, we show some tasks and instructions on how to apply our paradigm to the tasks.

For the tasks that have given targets (that is, Targeted Aspect-based Sentiment Classification, Conversational Aspect Sentiment Analysis), the first step is to adopt the domain- and target-constrained cloze test to generate the target-related candidate words that potentially provide the target’s property and background information. In this paper, we refer to these generated words as contextual target attributes since they are generated regarding the review context and contain the target’s attribute information. The second step is to design specific interaction graphs to integrate the beneficial attribute information into heterogeneous information interactions. 

For more general tasks that do not have given targets (that is, spoken language understanding, Dialog Sentiment Classification and Act Recognition), we first propose to use TF-IDF to extract the topic/key words, which can be regarded as ‘targets’. Then the domain- and target-constrained cloze test can be conducted to obtain the contextual target attributes, whose beneficial attribute information is then integrated into heterogeneous information interactions. More specifically, in dialog sentiment classification and act recognition, we can use the extracted ‘targets’ to model the topic transition in the dialog and the relation between the mentioned topics and speakers. We believe this information can provide some beneficial knowledge that PTLM cannot provide.

\section{Conclusion and Future Work}
In this work, we propose a new perspective to leverage the power of pre-trained language models for TSC: contextual target attributes, which are generated by our designed domain- 
 and target-constrained cloze test.
To exploit the attribute for TSC, we propose a heterogeneous information graph and a heterogeneous information gated graph convolutional network to capture the target sentiment clues via modeling the interactions between the target's background and property information, the syntactic information, and the contextual word correlative information.
Our method makes the first attempt to simultaneously leverage the merit of both masked language modeling and explicit target-context interactions.
Experiments show that we achieve new state-of-the-art performance.

Our approach makes the first attempt to simultaneously leverage the merits of masked language modeling (sufficient contextual knowledge) and fine-tuning method (explicit interactions achieved by upper-layer modules). Designed for TSC task, our approach has achieved significant improvements. In addition, our method provides further insights and can be generalized to other target-based or more general tasks in the future. 
For example, in dialog understanding tasks, we can use TF-IDF to extract the topic/key words which are regarded as targets. Then we can design some interaction graphs and corresponding networks to leverage these contextual extracted target attributes to model the topic transition in the dialog and the relation between the mentioned topics and speakers. This information is the potential to improve these tasks because it provides some beneficial knowledge that the PTLM cannot provide.

\section*{Acknowledgements}
This work was supported by Australian Research Council Grant DP200101328.
Bowen Xing and Ivor W. Tsang were also supported by A$^*$STAR Centre for Frontier AI Research.

\bibliographystyle{theapa}
\bibliography{ref}

\end{document}